\documentclass[letterpaper, 10 pt, conference]{ieeeconf}  

\IEEEoverridecommandlockouts                              

\usepackage{graphicx}
\usepackage{textcomp}

\title{\LARGE \bf
An Ensemble with Shared Representations Based on Convolutional Networks for Continually Learning Facial Expressions}

\author{Henrique Siqueira, Pablo Barros, Sven Magg and Stefan Wermter
\\\\\large{\textbf{Pre-print version of \cite{8594276}}}
\thanks{All of the authors are with Knowledge Technology, Department of Informatics, University of Hamburg, Vogt-Koelln-Str. 30, 22527 Hamburg, Germany {\tt\small\{siqueira, barros, magg, wermter\}@informatik.uni-hamburg.de}}}

\begin{document}

\maketitle
\thispagestyle{empty}
\pagestyle{empty}

\begin{abstract}
Social robots able to continually learn facial expressions could progressively improve their emotion recognition capability towards people interacting with them. Semi-supervised learning through ensemble predictions is an efficient strategy to leverage the high exposure of unlabelled facial expressions during human-robot interactions. Traditional ensemble-based systems, however, are composed of several independent classifiers leading to a high degree of redundancy, and unnecessary allocation of computational resources. In this paper, we proposed an ensemble based on convolutional networks where the early layers are strong low-level feature extractors, and their representations shared with an ensemble of convolutional branches. This results in a significant drop in redundancy of low-level features processing. Training in a semi-supervised setting, we show that our approach is able to continually learn facial expressions through ensemble predictions using unlabelled samples from different data distributions.
\end{abstract}

\section{Introduction}
Deep learning has made remarkable advances in the scientific community by improving state-of-the-art results in many domains such as speech recognition and image classification, and for the modern society by assisting us in our daily activities including web search and language translation \cite{Lecun2015}. The success of deep learning is usually attributed to the boost in computational power with parallel computing and the continuous release of large labelled datasets.

In the context of social affective robotics, however, gathering a large high-quality labelled dataset of emotions is tremendously difficult due to the intrinsic subjective perception of emotions \cite{Russell2003, Calbi2017}. Russell et al. \cite{Russell2003} suggest that emotion perception is more than a simple and universal decoding process, and may vary according to the internal emotional state of the receiver. Another factor is the influence of the context on emotion perception, as studied by Calbi et al. \cite{Calbi2017}. In their psychological experiments, they show evidence that a neutral facial expression cross-cut by a happy or fear content event is often perceived accordingly. This subjectivity on emotion perception reflects in high disagreement between annotators on categorizing emotional samples for a dataset, forcing authors to discard many samples or all of them of a given emotional category \cite{Nojavanasghari2016}.

Therefore, different learning strategies have been explored for enhancing emotional capabilities on social affective robots including learning through observation \cite{Feng2017}, and human rewards \cite{Churamani2017}. In the former, an autoencoder is used for learning appropriate facial expression responses by tracking facial landmarks from human-human conversations. In the latter, the authors have proposed a hybrid neural network architecture to teach facial expressions to a robot in an on-line fashion based on human feedbacks. However, a social robot would leverage the great exposure of continual information from the environment \cite{Parisi2018}, which could be very expensive through human feedback, or hard to evaluate the effectiveness in an auto-associative setting.

\begin{figure}[t]
	\centering
	\includegraphics[width=.48\textwidth]{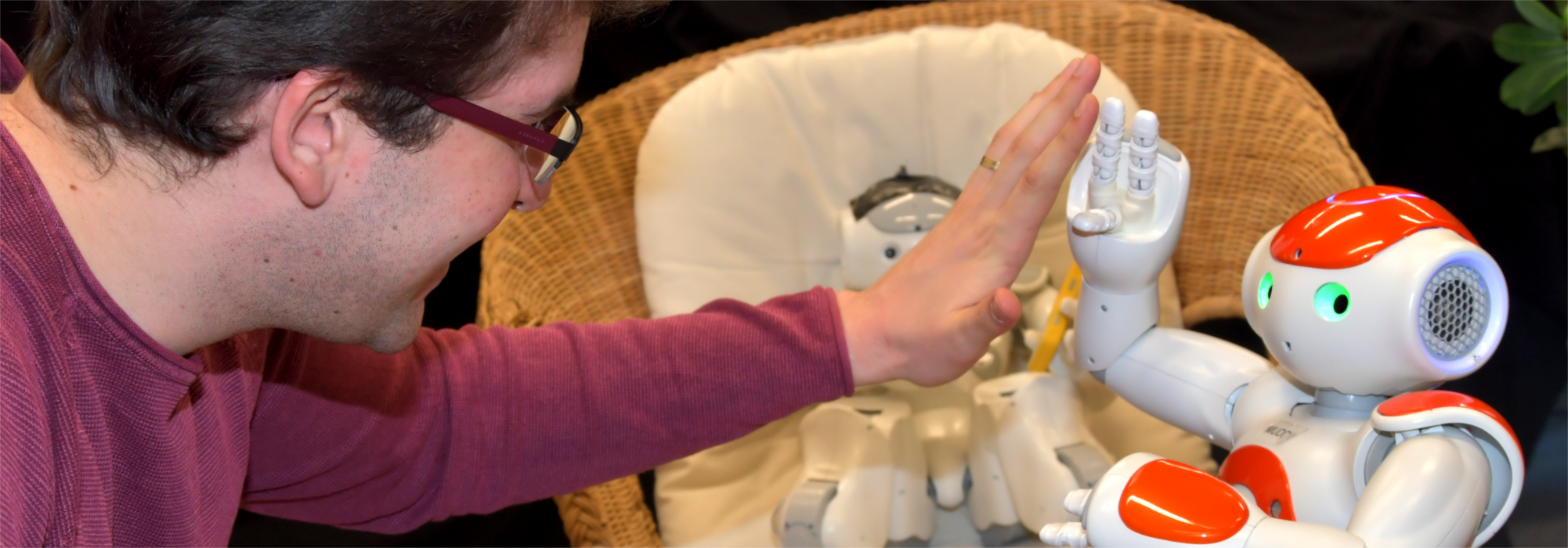}
	\caption{An HRI illustration where a NAO robot will continually learn facial expressions through close interactions using the proposed approach.}
	\label{fig:hri}
\end{figure}

A semi-supervised learning setting based on ensemble predictions could be a potential candidate, for instance, to exploit the great volume of unlabelled facial expressions exposed to a robot during human-robot interactions (HRIs) in order to improve its recognition capability, as illustrated in Figure \ref{fig:hri}. It is well-known that the generalization performance of an ensemble prediction from several classifiers is usually superior to the performance of a single classifier \cite{Freund1997}. This principle is adopted for training an ensemble composed of weak classifiers using a small amount of labelled data, and continually re-training the classifiers with plenty of unlabelled data through the ensemble predictions \cite{Blum1998}. As a result, a semi-supervised system can progressively improve its ability to solve a given task \cite{Mitchell2015}. However, the majority of ensemble-based systems are composed of several independent classifiers, which leads to an unnecessary great redundancy of low-level features processing, and consequently, high computational cost \cite{Malisiewicz2011}.

In this paper, we propose an ensemble based convolutional network for continually learning facial expressions through semi-supervised learning. Rather than traditional ensembles composed of several independent classifiers, the proposed ensemble consists of strong low-level feature extractors sharing convolutional branches for high-level feature extraction and classification designed as a single architecture. The use of shared representations results in a substantial drop in redundancy of low-level features processing, which reduces the computational cost of the ensemble, and makes the proposed approach suitable for social robots.

\section{Approach}
\begin{figure*}[t]
    \centering
    \includegraphics[width=0.78\textwidth]{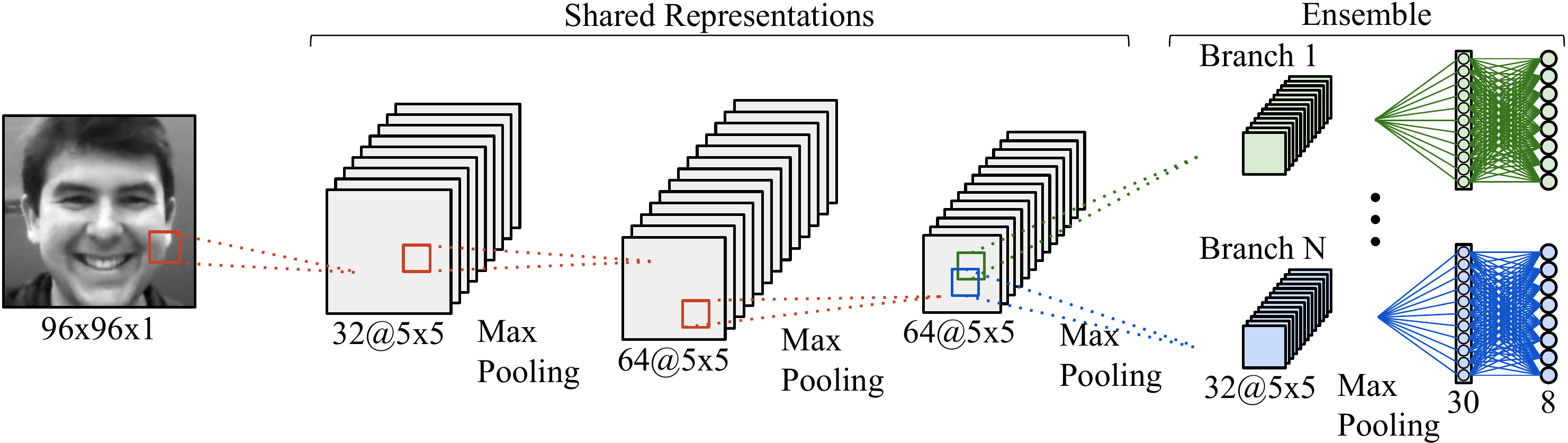}
    \caption{The proposed approach consists of convolutional layers for low-level feature extraction (on the left) shared with an ensemble of multiple convolutional branches for high-level feature extraction and classification (on the right).}
    \label{fig:approach}
\end{figure*}

Our approach relies on the hierarchical representational nature of convolutional neural networks (CNNs) \cite{Lecun2015} to design an ensemble architecture where early layers are strong low-level feature extractors, and their feature maps are shared with an ensemble of N weak independent classifiers designed as branches of convolutional layers followed by fully-connected layers for specific high-level feature extraction and classification, as shown in Figure \ref{fig:approach}. Using shared representational layers for low-level feature extraction rather than fully independent architectures results in a significant drop of redundant filters for detecting oriented lines, edges, colors, and textures, as well as simple parts of objects, especially for combinatorial-based methods like CNNs as investigated in our previous work \cite{Mousavi2016}. In contrast to ensemble methods that apply hand-engineered features and several basic classifiers like SVMs \cite{Malisiewicz2011}, our approach is not limited only to the generalization of different classification boundaries. Instead, it can adapt its representational space by re-training the shared layers, taking into account the learning signal from each high-level branch in the ensemble, leading to both novel features and novel classification boundaries.

Aiming to leverage unlabelled facial expressions that a social robot could gather during interactions with people for continually improving its automatic facial expression recognition system, the proposed architecture is trained in a semi-supervised learning setting. Firstly, we use a limited number of labelled samples for training the shared representational convolutional layers as strong low-level feature detectors, and each branch as a weak classifier of the ensemble. Subsequently, the trained network is ready to continually learn facial expressions by re-training the entire architecture using ensemble predictions from unlabelled samples. These two phases are detailed in the following sections.

\subsection{Training with Labelled Samples}
The first step of a semi-supervised learning setting based on ensemble predictions is to adopt a learning strategy that drives each weak classifier to develop specific representations from the labelled training data, such that the weak classifiers complement each other when exposed to unlabelled data. Otherwise, no additional knowledge can be acquired if all the weak classifiers succeed and fail on the same samples.

Therefore, we foster the development of different representations by separately training each of N convolutional branches with a different perspective of the labelled training data. For each training epoch, we iteratively train a given branch $b$ with a random portion from the labelled samples. Additionally, we also apply a random form of data augmentation before feeding the network to ensure that even if the same sample is presented to different convolutional branches, each branch processes a different version of such a sample.

\subsection{Re-training with Unlabelled Samples}
In this phase, the proposed architecture can be continually trained in parallel, i.e. as a single architecture, using ensemble predictions from unlabelled samples. The target output for a given unlabelled facial expression is obtained by a voting scheme, where each branch classification $c_{b}$ computes one vote of a voting vector $v$. Rather than taking the most voted category as target output, we apply the softmax function in order to re-train the network with a softer target probability distribution. According to Hinton et al. \cite{Hinton2015}, a soft target usually carries more distributional information about the task than a hard target output (e.g. one-hot output vector) when the data structure presents similarities.

\section{Experiments and Results}
Our objective is to propose a solution that is able to continually learn facial expression from unlabelled samples, therefore, in our experiments we are interested not only in reporting the overall generalization performance achieved by the proposed approach on a given dataset, but also on analysis of its learning behaviour when re-training with unlabelled facial expressions from unseen subjects. For instance, the latter experiment gives us evidence on how a social robot could progressively improve its facial expression recognition ability using images collected during human-robot interactions. In the following sections, we describe the datasets of facial expressions of emotion, the adopted methodology, the proposed architecture, the parameters used for training and re-training our approach, and finally, our results in multiple experimental conditions.

\subsection{Datasets}
\begin{figure}[h]
	\centering
	\includegraphics[width=.42\textwidth]{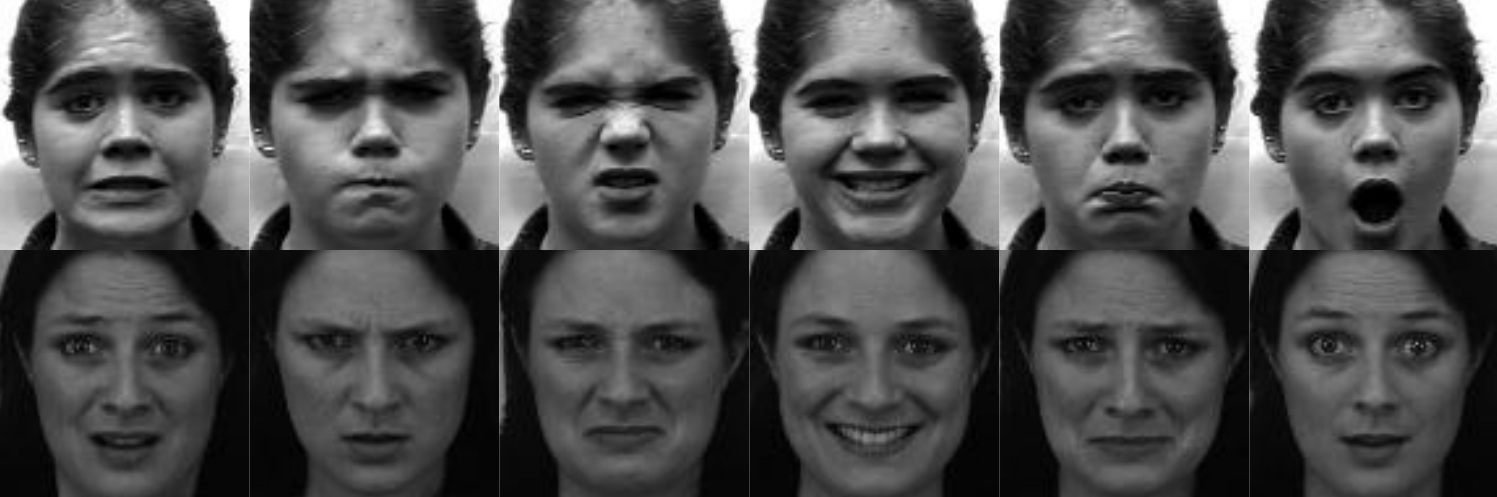}
	\caption{Fe, An, Di, Ha, Sa, and Su facial expressions from the CK+ (on top) and KDEF (on the bottom) datasets used in our experiments.}
	\label{fig:datasets}
\end{figure}

Our results are reported on two datasets of facial expressions of emotion exemplified in Figure \ref{fig:datasets}, the well-established Extended Cohn-Kanade (CK+) dataset \cite{Lucey2010} and the Karolinska Directed Emotional Faces (KDEF) dataset \cite{Lundqvist1998}. Both of the datasets are categorized in terms of discrete emotions based on the universal facial expressions according to Ekman \cite{Ekman1989}: Anger (An), Contempt (Co), Disgust (Di), Fear (Fe), Happiness (Ha), Sadness (Sa) and Surprise (Su), as well as neutral faces (Ne). In the KDEF dataset, however, the contempt category is not presented.

The CK+ dataset contains sequences of images from 123 subjects eliciting facial expressions, starting from a neutral face. In our experiments, we use the first frame as the neutral category and the last three frames as one of the seven emotional categories for training the network as a frame-based classifier. The KDEF dataset consists of static images from 70 subjects taken from different angles, however, only frontal face images are used in order to extend our experiments using CK+. Each subject provided two samples for each category.

\subsection{Methodology}
We adopt the subject-independent 10-fold cross-validation as our experimental methodology on the CK+ dataset. With this methodology, we ensure that the subjects selected to compose the test set are different from the ones in the validation and training sets. To populate each fold, initially, we sort the subjects in ascending order according to the subject's id provided by the dataset. Subsequently, with a step size of ten, all of the images from a given subject are sampled. As a result, each fold has around 12 subjects and 130.8 facial expression samples on average. In each trial, we pick Fold-$t$ for the test set, and Fold-$t+1$ for the validation set. Among the remaining eight folds, the first four folds are used as labelled samples and the last four folds as unlabelled samples. Thus, if Fold-9 is selected for testing, then Fold-10 is selected for validation, the folds from 1 to 4 are considered as labelled samples for the first training phase, and the folds from 5 to 8 are considered unlabelled samples for re-training the proposed approach with the ensemble predictions.

All of the samples from KDEF are reserved for investigating the learning behaviour when a different data distribution is used for re-training the proposed approach with unlabelled facial expressions. This experiment better represents realistic HRI scenarios, where a network is trained in advance using a public dataset and deployed in a real robot. Afterwards, the network should be able to improve facial expression recognition from facial expressions captured in different conditions, such as camera resolution, background, and so forth. Finally, the faces from both of the datasets are cropped by the Viola and Jones's algorithm for face detection \cite{Viola2004}, and re-scaled to 96 by 96 pixels.

\subsection{Architecture Details}
After assessing multiple architectures derived from the work of Khorrami et al. \cite{Khorrami2015} on the CK+ dataset to define a baseline, we fixed the overall configuration and only varied the number of branches that compose the ensemble in order to examine its influence on the generalization performance. As presented in Figure \ref{fig:approach}, the low-level shared representations layers are three convolutional layers with 32, 64 and 64 filters each, and a filter size of 5 by 5 and stride of 1. On top, each high-level representational branch is composed of one convolutional layer with 32 filters, filter size of 5 by 5 and stride of 1, followed by a fully-connected layer with 30 neurons and the output layer with 8 neurons. A max-pooling layer of 2 by 2 with a stride of 2 follows each convolutional layer. Batch-normalization is applied to regularize the convolutional layers, allowing the use of higher learning rates to increase training speed \cite{Ioffe2015}, which is a desirable property for on-line learning. As activation function, we adopted the hyperbolic tangent function except for the softmax function at the output layer.

\subsection{Training Details}
In the training phase using labelled data, our goal is to aid each convolutional branch to learn different and, more importantly, complementary representations from the training data. Therefore, we apply random forms of data augmentation to decrease the chance of the exact same input image being presented to different branches. The data augmentation consists of rescaling, translation, rotation, pixel intensity augmentation, and blurring. We train the network with the stochastic gradient descent (SGD) with a learning rate of 0.001, a decay of $10^{-5}$ and a momentum of 0.9.

In the re-training phase using unlabelled samples, the entire architecture is trained in parallel as a single architecture using SGD with a learning rate ten times smaller than in the previous phase and a decay of $2 \times 10^{-5}$. A learning rate equal or higher than in the previous phase had caused a critical forgetting problem, where facial expressions correctly recognized by the network became to be misclassified.

\subsection{Overall Generalization Performance on CK+}
First, the proposed approach is evaluated in terms of overall generalization performance by varying the number of branches. Table \ref{tab:rates} reports the average recognition rate percentage over ten trials of the 10-fold cross-validation on CK+ before and after the re-training phase.

\begin{table}
	\caption{Average recognition rate (\%) varying number of branches before and after re-training with unlabelled samples (US)}
	\center
	\begin{tabular}{| c | c | c | c |} \hline
		Branches		& 	Training 						& Validation		 				& 	Test								\\ \hline \hline
		1 			& 	99.67 \textpm \ 0.21				& 83.67 \textpm \ 5.45			&	82.38 \textpm \ 4.99				\\ \hline
		3			& 	99.67 \textpm \ 0.19				& 85.31 \textpm \ 4.86			&	85.09 \textpm \ 3.96				\\ \hline
		3 (US) 		& 	99.66 \textpm \ 0.28				& 86.69 \textpm \ 4.70			&	84.99 \textpm \ 3.40				\\ \hline
		5			& 	\textbf{99.81 \textpm \ 0.17}	& 86.23 \textpm \ 5.53			&	84.75 \textpm \ 5.72				\\ \hline
		5 (US)		& 	99.76 \textpm \ 0.18				& \textbf{87.92 \textpm \ 4.82}	&	\textbf{86.58 \textpm \ 5.62}	\\ \hline
	\end{tabular} 
	\label{tab:rates}
\end{table}

As expected, the lowest average recognition rate has been reached by the standard CNN with one branch, confirming that an ensemble of classifiers generally leads to a better generalization performance than a single classifier. The re-training with unlabelled samples has increased the average recognition rates on both of the validation and test sets for an ensemble with five branches to 87.92\% and 86.85\%, respectively, but with a relatively high standard deviation. Therefore, to verify if there is a statistical difference after re-training the network with unlabelled samples, we performed a paired t-test which has given us a p-value of 0.1.

\begin{table*}
	\caption{Recognition rate (\%) over folds for the ensemble of 5 branches before and after re-training with unlabelled samples (US)}
	\center
	\begin{tabular}{| l | c | c | c | c | c | c | c | c | c | c |} \hline
		Folds			&	1				&	2		&	3		&	4		&	5			&		6				&	7		&	8				&	9				&	10		\\ \hline \hline
		5 Branches (US)	&	85.60			&	82.69	&	86.25	&	82.14	&	89.16			&	77.02			&	86.20	&	98.28			&	93.18			&	85.22	\\ \hline
		5 Branches 		&	82.57			&	83.33	&	86.25	&	82.14	&	87.50			&	70.27			&	88.79	&	93.10			&	87.12			&	86.36	\\ \hline
		Difference		&	\textbf{+3.03}	&	-0.64	&	0.00		&	0.00		&	\textbf{+1.66}	&	\textbf{+6.75}	&	-2.58	&	\textbf{+5.17}	&	\textbf{+6.06}	&	-1.13	\\ \hline
	\end{tabular} 
	\label{tab:folds}
\end{table*}

Table \ref{tab:folds} shows the recognition rate obtained by the ensemble predictions for each trial, from which we can identify individual cases of improvement and decrement. In five of ten trials, re-training the network with unlabelled data has increased the recognition rate on the test set by more than one percent. Among them, three show a significant improvement of more than five percent. In the unsuccessful cases, two of ten trials have decreased the recognition rate by more than one percent. This analysis gives evidence that the use of unlabelled facial expressions for re-training the proposed architecture could lead to significant improvement on accuracy, which is suitable for a social robot that is continually interacting with people during social events.

\subsection{Learning Behaviour Analysis}
\begin{figure}[h]
	\centering
	\includegraphics[width=.43\textwidth]{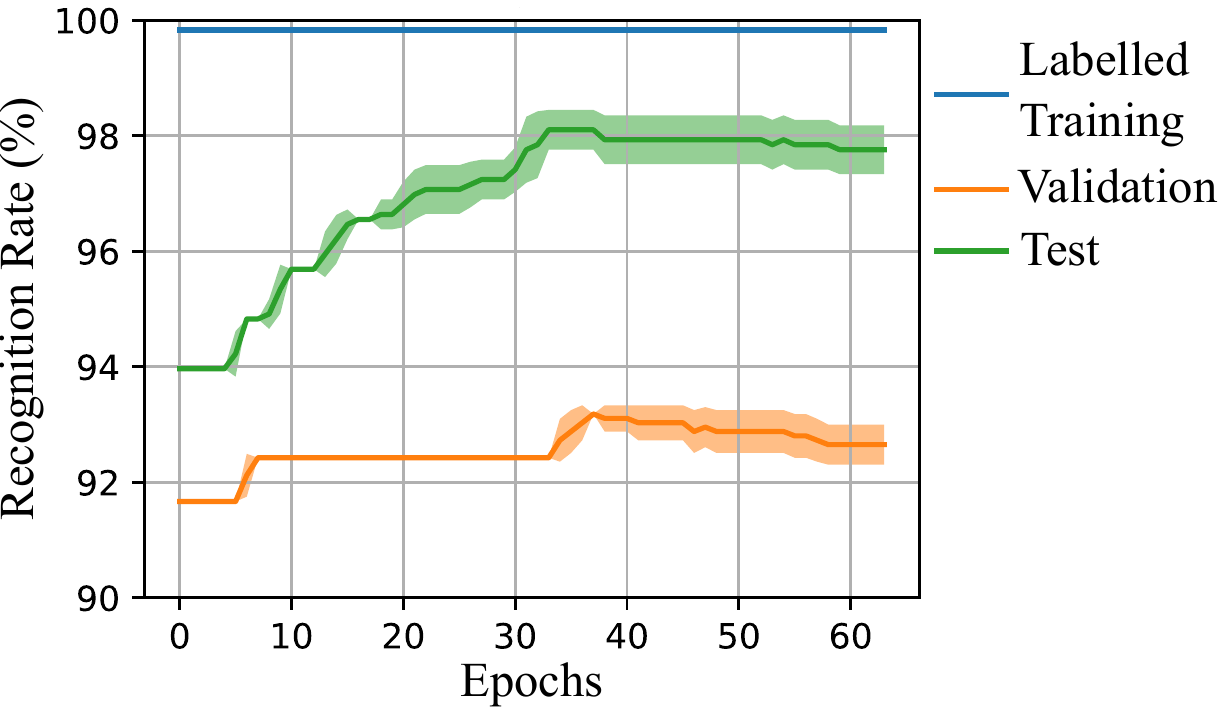}
	\caption{\textbf{Fold-8 on CK+.} Average recognition rate curves during the re-training phase in the best learning behaviour case.}
	\label{fig:conv8}
\end{figure}

To analyse the ensemble's impact on learning, we repeated the experiment for the best (Fold-8) and worst (Fold-7) cases over the ten trials from the previous experiment. In the best trial, the network has presented a significant improvement on generalization performance after the re-training phase and the highest recognition rate on the test set. Figure \ref{fig:conv8} shows the average recognition rate curves (i.e. the labelled training, validation and test sets) after 10 runs using Fold-8 for the test on CK+ during the re-training phase with ensemble predictions. Therefore, the origin of the graph corresponds to the recognition rate of the ensemble prediction immediately after the training phase with labelled data.

\begin{figure*}[h]
    \centering
    \includegraphics[width=1.0\textwidth]{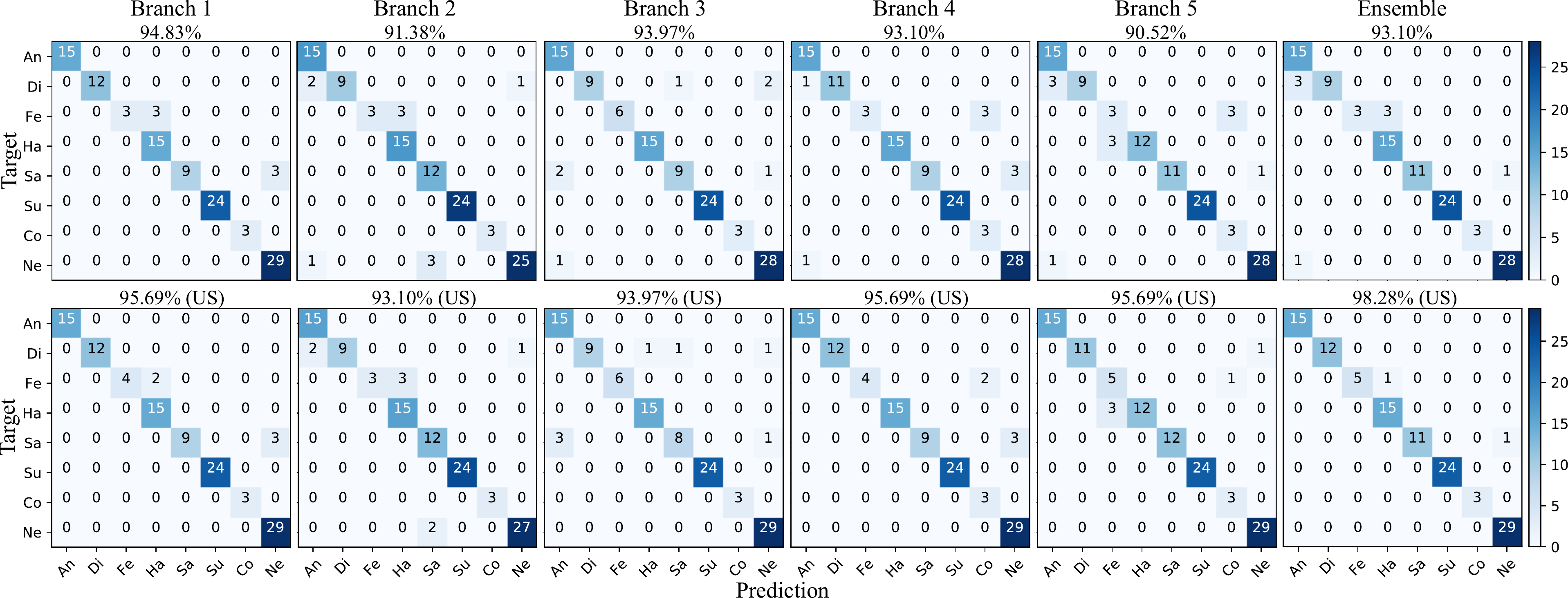}
    \caption{Confusion matrices with the recognition rates for each branch prediction in the best learning behaviour case (Fold-8) using the ensemble of 5 branches. On top, results before re-training the ensemble, and on the bottom after re-training with unlabelled samples (US).}
    \label{fig:cm8branches}
\end{figure*}

Observe that the generalization performance on CK+ has improved for both the validation and test sets, and the forgetting problem has not affected the network in this fold. Moreover, the learning behaviour is consistent over the 10 trials, by keeping the same improvement behaviour with a low standard deviation. Note that, all of the branches have improved the generalization performance on CK+ through the ensemble prediction, as shown by the confusion matrix for each branch and the ensemble classification in Figure \ref{fig:cm8branches}.

Through the analysis of the confusion matrices, we are able to investigate what promotes the improvement on generalization performance and how each branch's expertise complements each other. Let's consider branch 3 before the re-training phase: it has achieved the best recognition performance for the fear expressions due to its high true-positive and low false-positive rates for the classification of such expression. In other words, branch 3 correctly classifies six fear facial expressions from a total of six on the test set and misclassified no other expressions as fear. The same explanation is valid for branch 2, which is the best branch for sadness compared to the other branches of the ensemble.

These findings suggest that the improvement in the overall generalization performance on CK+, in this case, has been caused by the high complementary expertise between these branches. While a given branch is a strong classifier for a given category such as fear but weaker for sadness, another branch can provide the complementary expertise. This yields to an improvement on generalization performance on Fold-8 by re-training the network with additional unlabelled data through ensemble predictions. Branch 5 becomes a stronger classifier for fear facial expressions, likely throwing a correct teaching signal from the branch 3 on such expression. Finally, the overall generalization performance has improved from 93.10\% to 98.28\% after the re-training phase.

\begin{figure}[h]
	\centering
	\includegraphics[width=.43\textwidth]{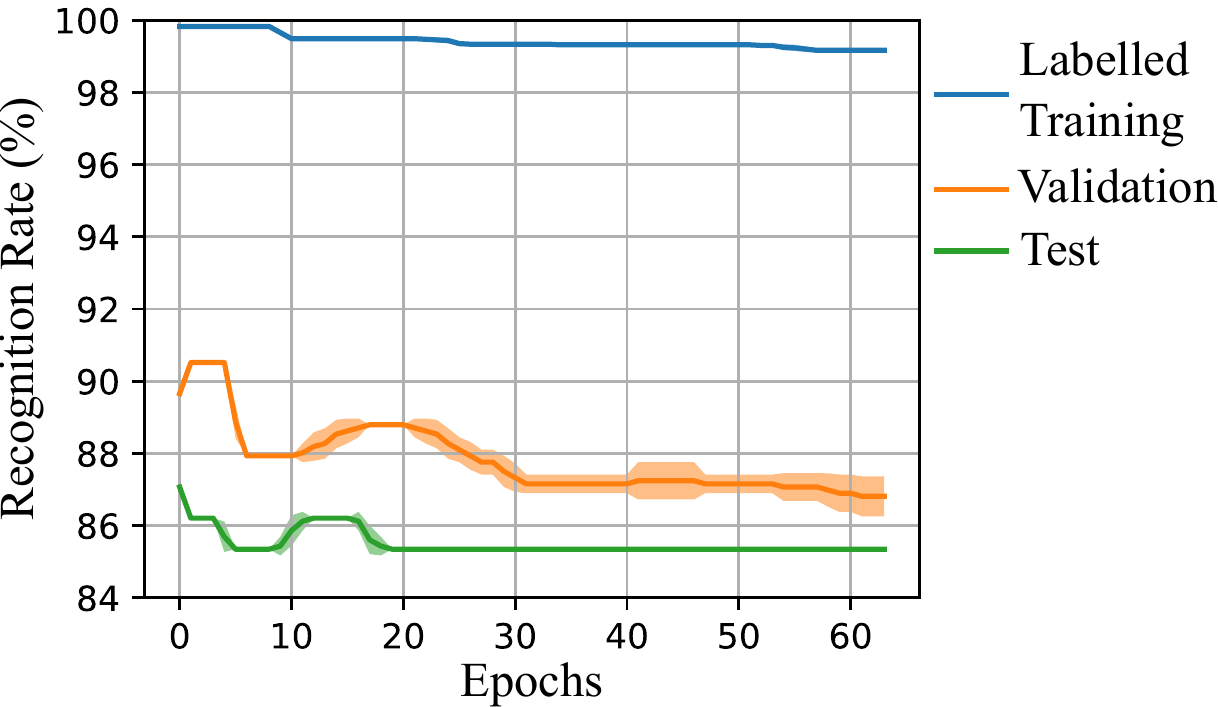}
	\caption{\textbf{Fold-7 on CK+.} Average recognition rate curves during the re-training phase in the worst learning behaviour case.}
	\label{fig:conv7}
\end{figure}

In the worst trial (using Fold-7 for the test set), both of the validation and test sets have decreased around 2\% on average, and the forgetting problem subtly affected the network during the re-training phase, as shown in Figure \ref{fig:conv7}. Despite such undesirable behaviour, after 20 epochs the generalization performance on the test set stops degenerating and stays in a stationary condition.

\begin{figure}[h]
	\centering
	\includegraphics[width=.43\textwidth]{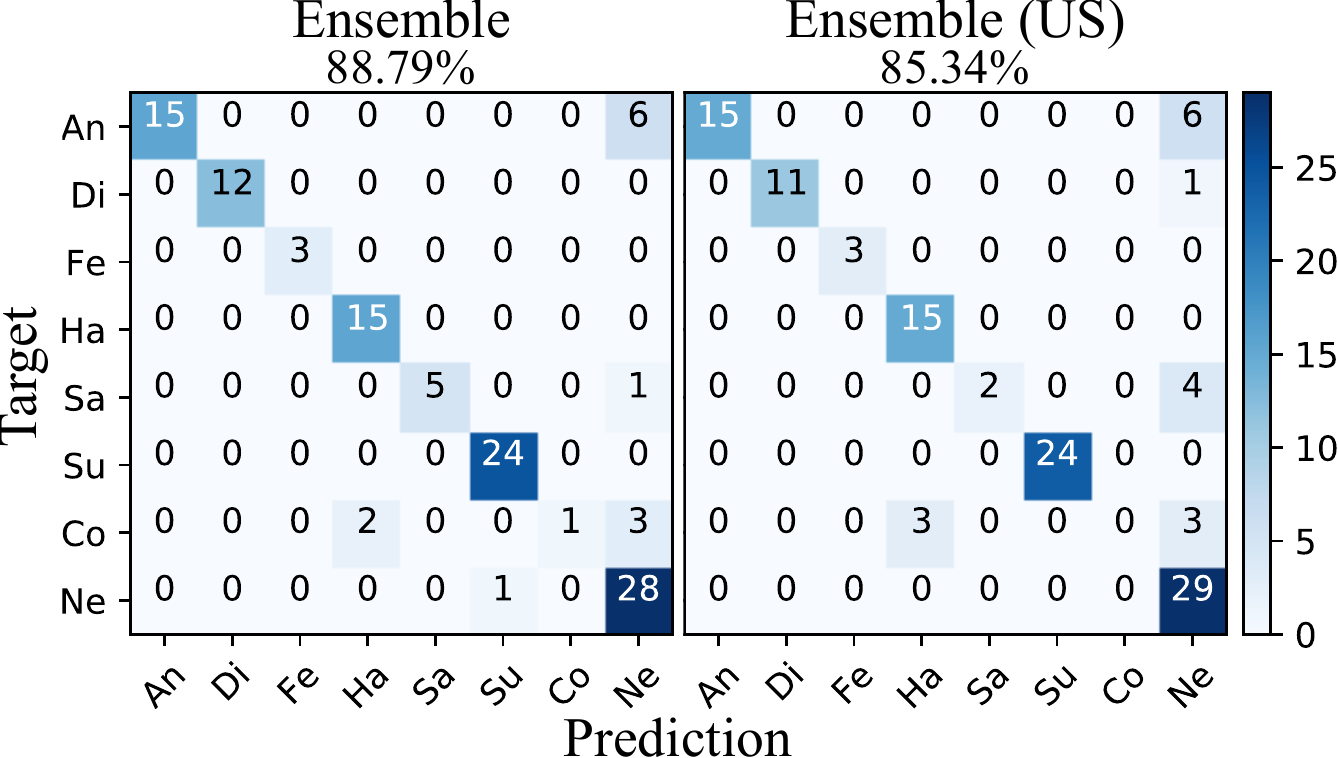}
	\caption{Confusion matrices with the recognition rates in the worst learning behaviour case (Fold-7) using the ensemble of 5 branches.}
	\label{fig:cm7}
\end{figure}

Examining the confusion matrix from the ensemble predictions before the re-training phase in Figure \ref{fig:cm7}, the significantly high number of facial expressions misclassified as neutral indicates a strong bias of the ensemble. The bias for the neutral category becomes even stronger after the re-training phase, where strong branches for disgust and sad facial expressions might have been encouraged to wrongly classify such samples as neutral.

\subsection{Continually Learning Facial Expressions on KDEF}
In general, re-training the proposed architecture with unlabelled data from unseen subjects has shown to be beneficial by improving overall generalization performance on CK+, however, it is also necessary to investigate the learning behaviour using facial expressions from a different data distribution. It is especially needed for facial expression recognition since people usually express themselves in many different ways \cite{Barrett2006}, as well as for evaluating the proposed approach using images recorded in a different setting, which happens in real HRI where images are captured by the robot's camera. These differences can be observed in Figure \ref{fig:datasets}.

\begin{figure}[h]
	\centering
	\includegraphics[width=.49\textwidth]{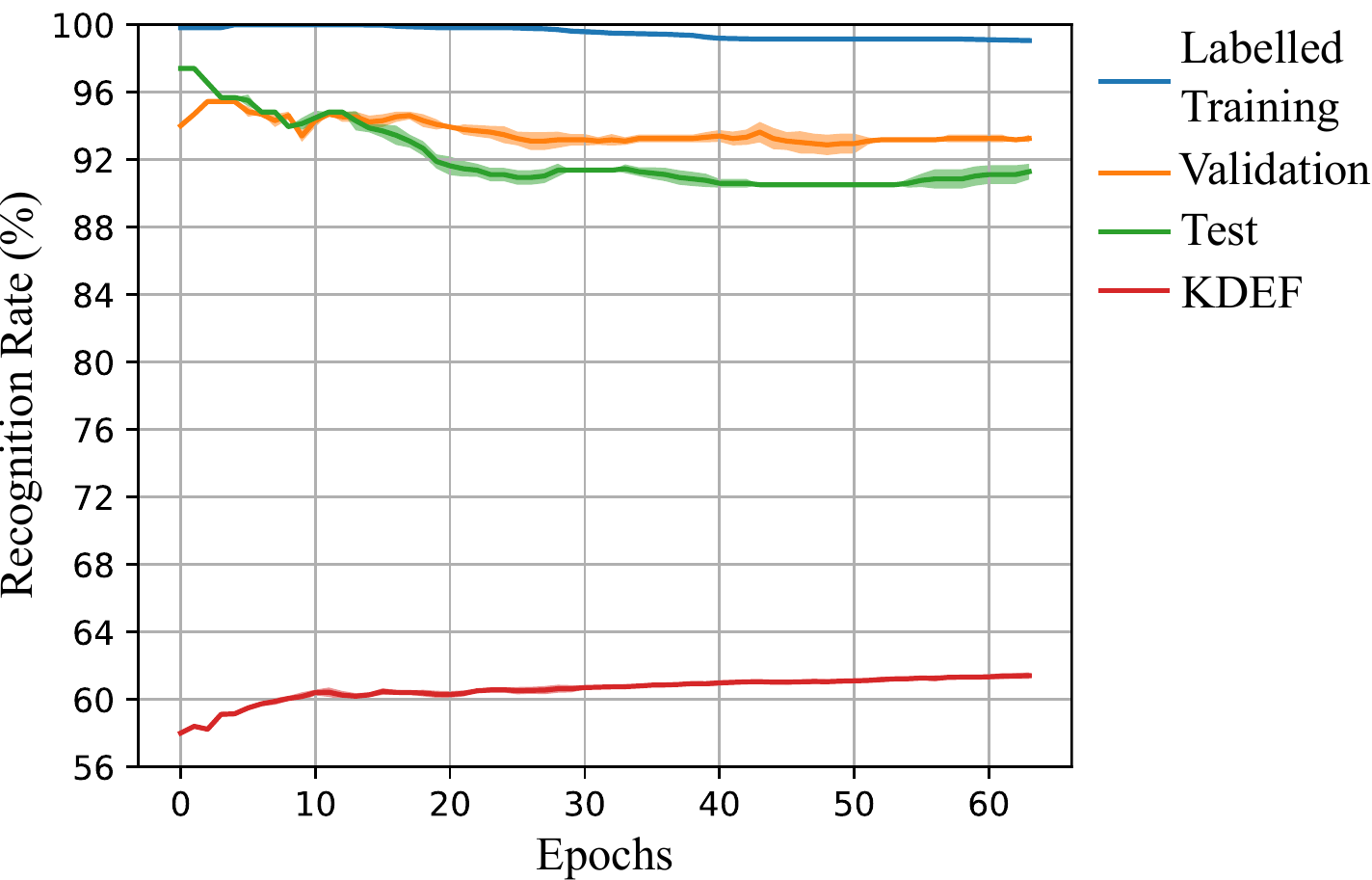}
	\caption{Average recognition rate curves on CK+ and KDEF during the re-training phase using the KDEF dataset as unlabelled samples.}
	\label{fig:convkdfe}
\end{figure}

Figure \ref{fig:convkdfe} shows the average recognition rate as a function of epoch using facial expressions from the KDEF dataset for re-training the ensemble. The large recognition gap between CK+ and KDEF is expected when a trained network is tested in a different data distribution. Nevertheless, the learning behaviour on KDEF is stable and consistent with a continually boost on generalization performance over epochs on KDEF, low standard deviation after ten trials, and an increment of around 4\% on average at the end. Finally, although the forgetting problem affected the network for expressions from CK+, our major goal is to increase recognition performance towards people who interact with a social robot instead of keeping generalization performance on a specific dataset, which is simulated in this experiment by using a dataset with a different data distribution.

\section{Conclusions and Future Work}
A social robot capable of continually learning facial expressions from its previous knowledge using unlabelled facial expressions collected during daily HRIs could enhance its emotional responses towards people who share the same environment with such a robot. In order to address this problem, we proposed an ensemble approach with shared representations based on convolutional networks. Using shared representational layers, we decreased the number of redundant low-level overall processing compared to traditional ensembles composed of independent classifiers, which make our ensemble approach suitable for robotic platforms with limited computational resources.

By re-training the proposed approach with unlabelled samples from unseen people, we improved the generalization performance on the CK+ and KDEF datasets. Analysing the confusion matrices of the ensemble, we demonstrated that the more divergent the branches' expertise, the higher is the improvement on generalization performance. However, at least one branch has to succeed in the prediction of a given sample for an effective learning. In psychology, this is called desirable difficult examples, where a learner has sufficient knowledge to succeed. As a result, the accomplished task stimulates learning, comprehension, and remembering \cite{Bjotk2011}.

As future work, strategies for increasing the ensemble diversity will be explored in order to prevent degeneration on generalization performance. Our promising results motivate us to extend the proposed architecture to deal with multimodal inputs including audio and gestures for two main reasons: (i) the processing of multimodal inputs usually leads to improvement on emotional expression recognition \cite{Poria2017}, and (ii) a different view from the same output task can assist the learning process by constraining the problem as suggested by Mitchell et al. \cite{Mitchell2015}. In the accompanying video (\textit{https://goo.gl/qFXhcK}), we demonstrate how our approach can be used to improve recognition performance under real-world conditions. However, the progressive learning behaviour of the proposed approach will also be investigated in an HRI scenario using a NAO robot as illustrated in Figure \ref{fig:hri}, where it should give the most appropriate response according to facial expressions, and continually improve its emotion recognition capability using unlabelled facial expressions captured from close communications with humans.

\section{Acknowledgement}
This work has received funding from the European Union under the SOCRATES project (No. 721619), and the German Research Foundation under the CML project (TRR 169). We would like to thank Erik Strahl for his valuable contributions that improved the quality of this paper.

\bibliographystyle{IEEEtran}
\bibliography{refs}
\end{document}